\documentclass[letter,conference]{IEEEtran}
\IEEEoverridecommandlockouts
\usepackage{cite}
\usepackage{amsmath,amssymb,amsfonts}
\usepackage{graphicx}
\usepackage{placeins}
\usepackage{textcomp}
\usepackage{xcolor}
\usepackage{mathtools}
\usepackage{amsmath}
\usepackage{flafter}
\usepackage{pifont}
\usepackage{tabularx}
\usepackage{color, colortbl}
\usepackage{tabularray}
\usepackage{lipsum}
\usepackage{graphicx}
\definecolor{lime}{rgb}{0.88,2,10}
\usepackage[left=0.705in, right=0.705in, top=0.71in, bottom=0.7in]{geometry}
\usepackage[linesnumbered,ruled,vlined]{algorithm2e}
\renewcommand{\baselinestretch}{.998}

\SetKwInput{kwInit}{Initialize}
\SetKwComment{Comment}{$\triangleright$\ }{}

\usepackage{subcaption}
\usepackage{wrapfig}

\usepackage[linesnumbered,ruled,vlined]{algorithm2e}
\usepackage{xpatch}
\usepackage{url}
\usepackage{multirow}
\usepackage{multicol}
\usepackage{wrapfig}
\usepackage{algorithmic}
\usepackage[export]{adjustbox}
\usepackage{nomencl}
\usepackage{siunitx}
\usepackage{blindtext}
\usepackage[T1]{fontenc}
\usepackage[spaces,hyphens]{xurl}
\usepackage{float}
\usepackage{stfloats} 

\SetAlFnt{\small}

\SetKwComment{Comment}{$\triangleright$\ }{}

\usepackage[utf8]{inputenc}
\usepackage{enumitem}

\SetKwInput{kwInit}{Initialize}
\def\BibTeX{{\rm B\kern-.05em{\sc i\kern-.025em b}\kern-.08em
    T\kern-.1667em\lower.7ex\hbox{E}\kern-.125emX}}

\usepackage[numbers,sort&compress]{natbib}
\usepackage[font={small}]{caption} 

\newcommand{\fref}[1]{Fig.~\ref{#1}}
\newcommand{\tref}[1]{Table~\ref{#1}}
\newcommand{\sref}[1]{Section~\ref{#1}}
\usepackage[misc]{ifsym}

\usepackage{fancyhdr}
\fancypagestyle{mystyle}{
    \chead{\large 2023 IEEE International Conference on E-health Networking, Application \& Services (Healthcom 2023)}}

\begin{document}

\title{Diagnosing Alzheimer's Disease using Early-Late Multimodal Data Fusion with Jacobian Maps}
\author{Yasmine Mustafa, Tie Luo$^*$\\ {Computer Science Department, Missouri University of Science and Technology, USA}\\ Email:  \{yam64, tluo\}@mst.edu\thanks{$^*$ Corresponding author.}}

\maketitle
\thispagestyle{mystyle}
\begin{abstract}
Alzheimer's disease (AD) is a prevalent and debilitating neurodegenerative disorder impacting a large aging population. Detecting AD in all its presymptomatic and symptomatic stages is crucial for early intervention and treatment. An active research direction is to explore machine learning methods that harness multimodal data fusion to outperform human inspection of medical scans. However, existing multimodal fusion models have limitations, including redundant computation, complex architecture, and simplistic handling of missing data. Moreover, the preprocessing pipelines of medical scans remain inadequately detailed and are seldom optimized for individual subjects. In this paper, we propose an efficient early-late fusion (ELF) approach, which leverages a convolutional neural network for automated feature extraction and random forests for their competitive performance on small datasets. Additionally, we introduce a robust preprocessing pipeline that adapts to the unique characteristics of individual subjects and makes use of whole brain images rather than slices or patches. Moreover, to tackle the challenge of detecting subtle changes in brain volume, we transform images into the Jacobian domain (JD) to enhance both accuracy and robustness in our classification. Using MRI and CT images from the OASIS-3 dataset, our experiments demonstrate the effectiveness of the ELF approach in classifying AD into four stages with an accuracy of 97.19\%. 
\end{abstract}


\section{Introduction}
Alzheimer's disease (AD) is one of the most prevalent and serious degenerative diseases of the brain, affecting 1 out of 9 people over the age of 65 years \cite{alzheimer20132013}. AD is not a mere continuum; rather, it encompasses a silent progressive nature for several years before the onset of mild cognitive impairment (MCI) when the subject experiences memory or thinking problems. Subsequently, 10\%-15\% of individuals with MCI progress to AD each year \cite{petersen2009mild}. AD patients have a poor quality of life, memory loss, and a constant need for support. It is estimated that about 1.2\% of the world's population will develop AD by 2046. These statistics necessitate the development of biomarkers to diagnose AD in its different stages. However, distinguishing subtle pattern changes in the brain through manual analysis of brain imaging scans, such as magnetic resonance imaging (MRI) or computed tomography (CT) scans, can be challenging and cumbersome due to the large uncertainty in the diagnosis process \cite{salami2022designing, johnson2012brain, huang2020fusion}.

{\bf Related Work.} In response, a large body of research has emerged focusing on neuroimaging-based computer-aided classification of AD and its prodromal stage, MCI \cite{rathore2017review}. Neuroscience institutions have been developing large-scale datasets of various modalities to foster research on AD classification and dementia diagnosis. By harnessing and combining the unique information conveyed by each modality about the brain, machine learning with multimodal data has demonstrated promising performance compared to relying solely on a single modality \cite{venugopalan2021multimodal}. However, to date, effective multimodal data fusion in machine learning remains a daunting challenge, which involves fusion without losing or compromising valuable information and without disrupting the complementary relationship between different imaging scans. 

Strategies of fusion in the literature can be divided into three types \cite{huang2020fusion}: Early fusion, also known as {\em feature-level fusion}, involves combining multimodal data by joining their features in a vector that is then fed into a machine learning model. Joint fusion refers to an intermediate fusion that joins the feature representations learned from one modality at {\em intermediate layers} of a neural network with feature representations learned from other modalities. Late fusion employs {\em decision-level fusion}, where a separate model is trained for each modality and all of the models' predictions are then combined to make a final decision. 

Different fusion strategies have been explored in previous studies. For instance, Liu et al. \cite{liu2018multi} introduced a cascaded 3D convolutional neural network (CNN) which involves pretraining two 3D CNNs over local patches extracted from 3D MRI and 3D positron emission tomography (PET) images, respectively. The separately learned features are then fused by a regular (2D) CNN, whose output feature maps are sent to a final fully connected layer.
Similarly, Abdelaziz et al. \cite{abdelaziz2021alzheimer} used T1-weighted (T1w) MRI, PET, and single nucleotide polymorphisms (SNPs) to train three different CNNs respectively, before fusing the three CNNs' features into a fully connected layer. On the other hand, Li et al. \cite{li2019early} applied early fusion of MRI data and clinical assessments to predict the progression of AD in MCI patients using recurrent neural networks (RNNs).  Qiu et al.  \cite{qiu2018fusion} also applied early fusion of MRI and two clinical assessment data, Mini-Mental State Examination (MMSE) and logical memory (LM), where MRI images were divided into three slices and fed individually to three VGG-11 models, respectively, and majority voting was then used to aggregate the three VGG-11 models. As for the MMSE and LM, they were also fed separately to two models and the final result was obtained by combining the aggregation of the VGG-11 models and the MMSE and LM models. Despite the various fusion strategies explored in the literature, an effective fusion strategy that consistently achieves optimal performance across domains remains an open problem \cite{huang2020fusion}. 

Besides data fusion, multimodal data also poses other challenges such as appropriate model selection. Researchers used to choose shallow models like support vector machines (SVM) and random forest; for example, Bi et al. created brain region-gene pairs using structural MRI and gene data and then clustered them using clustering evolutionary
random forest (CERF) \cite{bi2020multimodal}. However, recent studies show that deep models typically perform better in Alzheimer's diagnosis \cite{venugopalan2021multimodal}. 

{\bf Gaps in the Literature.} Although deep neural networks have excelled in many problem domains, their typical demand for large datasets presents a challenge to neuroimaging datasets which are often relatively small. This motivated studies like \cite{salami2022designing} to use pretrained networks like ResNet, and other studies that employ data augmentation, feature fusion, or decision fusion. However, these only partially solve or mitigate the problem due to the big difference between data supply and demand.
Another challenge is posed by missing modalities of some subjects, which leads to a significant loss of valuable information during data fusion as many studies simply eliminate subjects with missing modalities. Even though some studies have used techniques like linear interpolation to address missing data \cite{abdelaziz2021alzheimer}, there is still large room for technique (re)design and performance improvement.

Another limitation in existing work based on joint fusion is {\em redundant computation} caused by the adoption of the same (sub)network architecture for all modalities. That approach forgoes the potential benefits of using a different model tailored specifically to each modality. On the other hand, having different initial models for each modality can be unwieldy, especially when dealing with a considerable number of modalities. Therefore, it is a challenging dilemma between settling on a single architecture or multiple modality-specific architectures, due to the variation in data-dependent factors such as preprocessing and data attributes.

One more limitation observed in joint fusion is its inability to make predictions when not all the modalities are present, which, however, is common in the real world \cite{huang2020fusion}. On the other hand, late fusion can handle missing data because models are trained independently; however, it lacks interactions between features extracted from different modalities \cite{yoo2019deep}. One more advantage of early and late fusions over joint fusion is that they do not require intricate design but can yield competitive performance when properly used. 

{\bf Our Work and Contribution.}
In this paper, we propose a pragmatic early-late fusion (ELF) approach that addresses the following challenges: 1) handling subjects with missing modalities, 2) managing small datasets, and 3) mitigating redundant networks and engineering complexity, as commonly encountered in joint fusion methods. Our ELF approach amalgamates the strength of both deep and shallow models, comprising (i) a lightweight CNN that works in conjunction with early-fused Jacobian determinant features, (ii) random forest (RF) models operating alongside pretrained ResNets, and (iii) a late fusion mechanism that combines predictions from both the CNN and RF models.
%
Specifically, this paper makes the following contributions:
\begin{itemize}[leftmargin=*]
    \item Our proposed ELF approach provides a multimodal data fusion solution that enhances automated AD diagnosis with improved performance, scalability, and ease of implementation.
    
   \item We introduce Jacobian maps into our multimodal model and thus enable capturing subtle brain volume changes and provide more informative representations for feature learning. 
    
    \item We account for individual subject variations in brain shape by employing per-subject adaptation registration,\footnote{Registration in the context of medical imaging is a process of aligning and transferring different medical images to a common coordinate system.} thus providing a {\em personalized} diagnostic approach.

   \item This work represents the first effort of applying hot deck imputation (HDI) based on kurtosis and skewness to address missing modalities in the context of multimodal data fusion.
   
    \item Our model allows for training over whole-brain scan images, which preserves spatial correlations and improves accuracy as opposed to existing patch or slice-based approaches.

    \item Unlike existing studies which explore only binary classification for AD (i.e., positive/negative), ELF achieves {\em four-stage classification} with an impressive accuracy of 97.19\%, contributing to {\em precision medicine}.
\end{itemize}


\begin{figure*}
    \scalebox{.7}{
\centering \subfloat[\small{Raw Images}]{\includegraphics[width=0.18\linewidth]{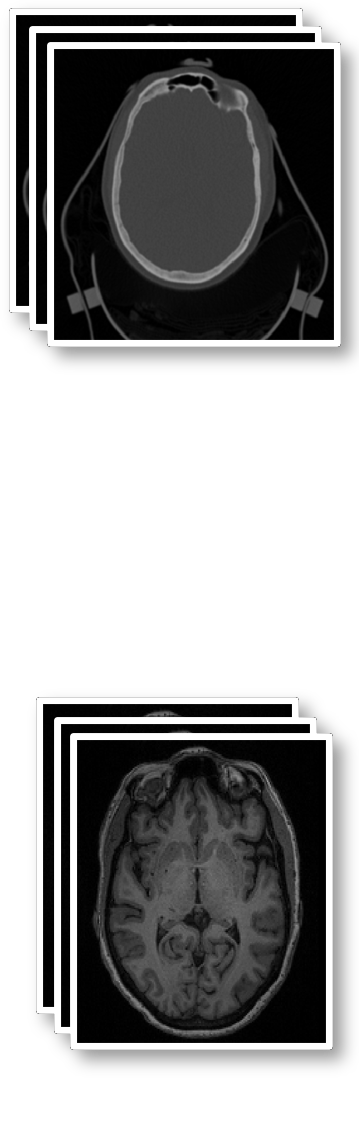}}
 \hspace{0.2cm}\subfloat[\centering\small{Contrast stretching/Bias Correction}]{\includegraphics[width=0.23\linewidth]{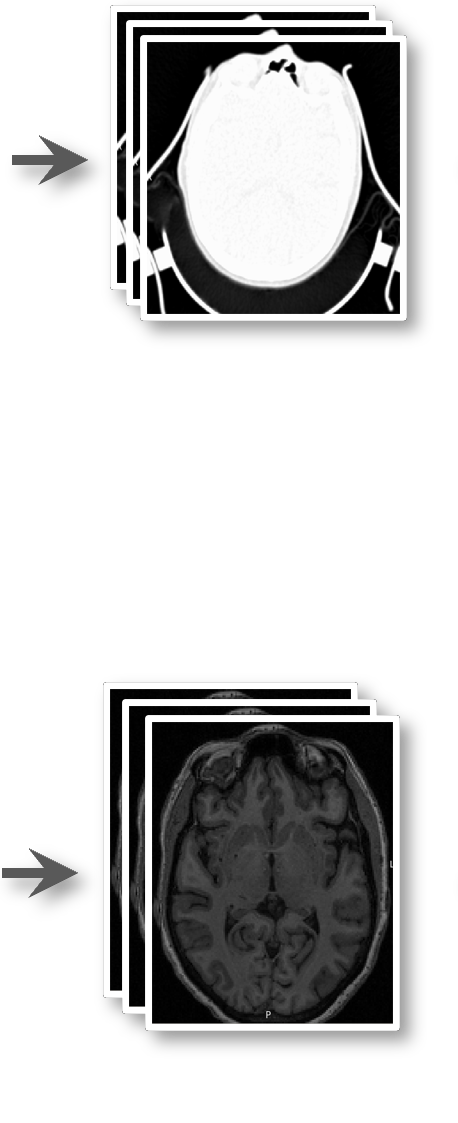}}
    \hspace{0.1cm}\subfloat[\small{BET}]{\includegraphics[width=0.225\linewidth]{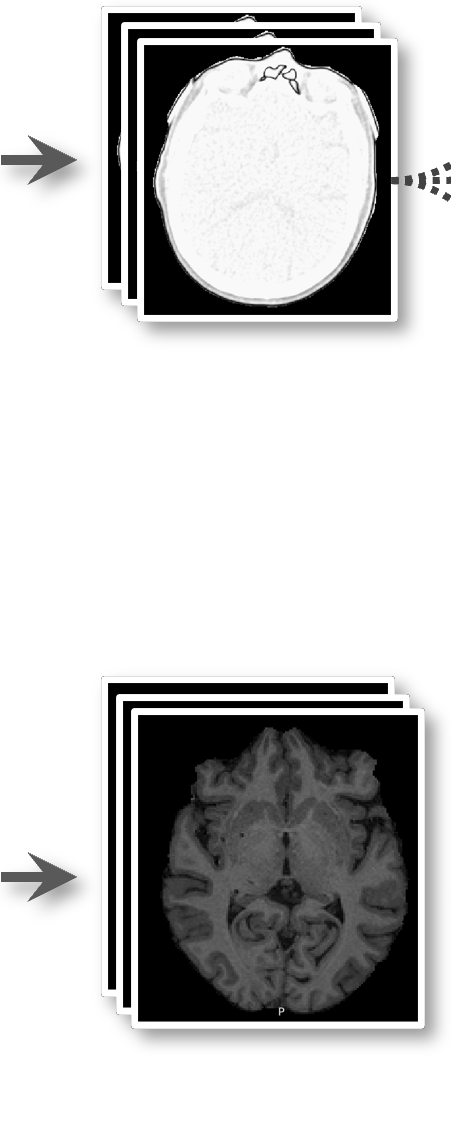}}
  \subfloat[\small{Multi-stage Registration}]{\includegraphics[width=0.529\linewidth]{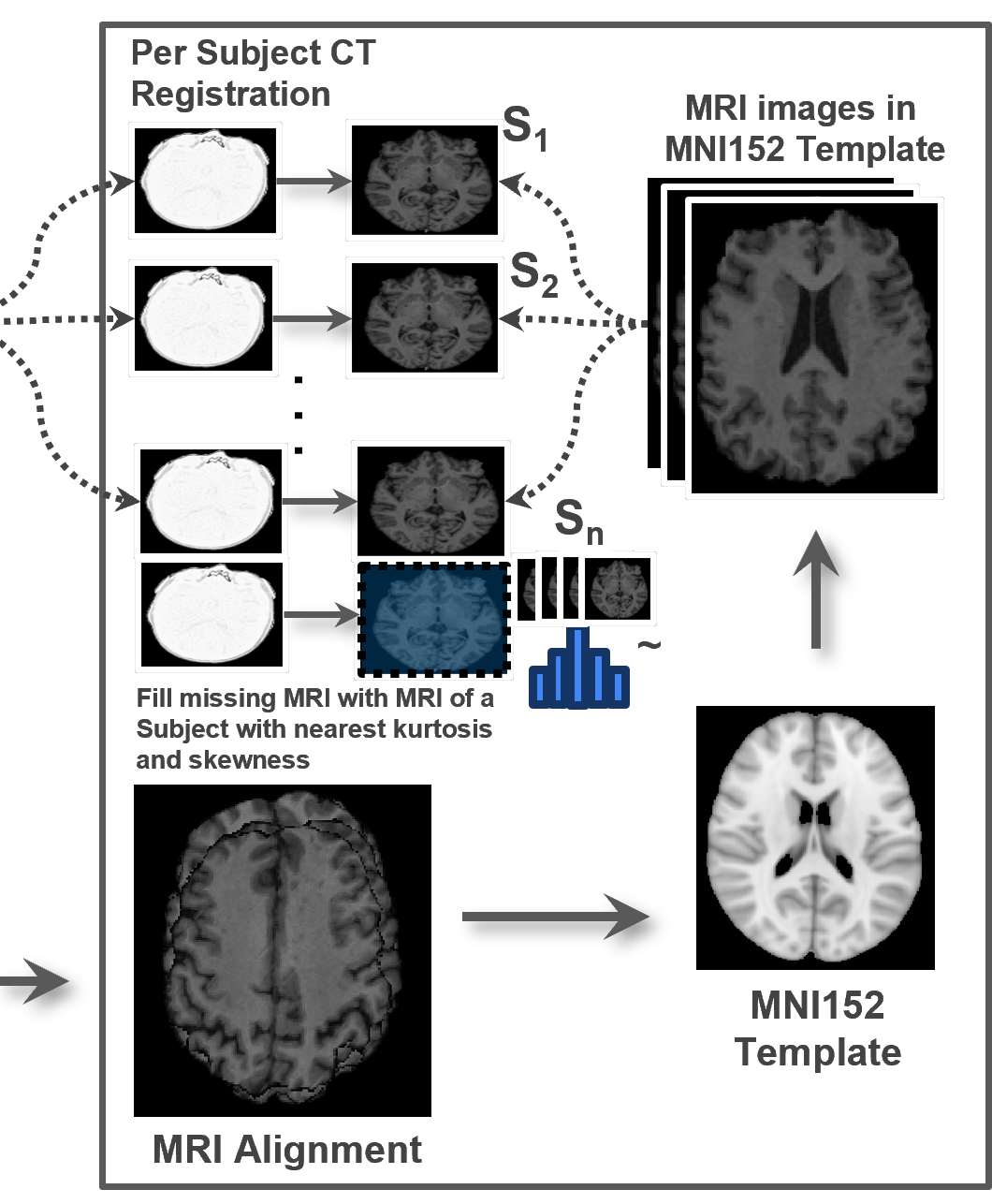}}
    \hspace{0.2cm}\subfloat[\small{Jacobian Maps}]{\includegraphics[width=0.22\linewidth]{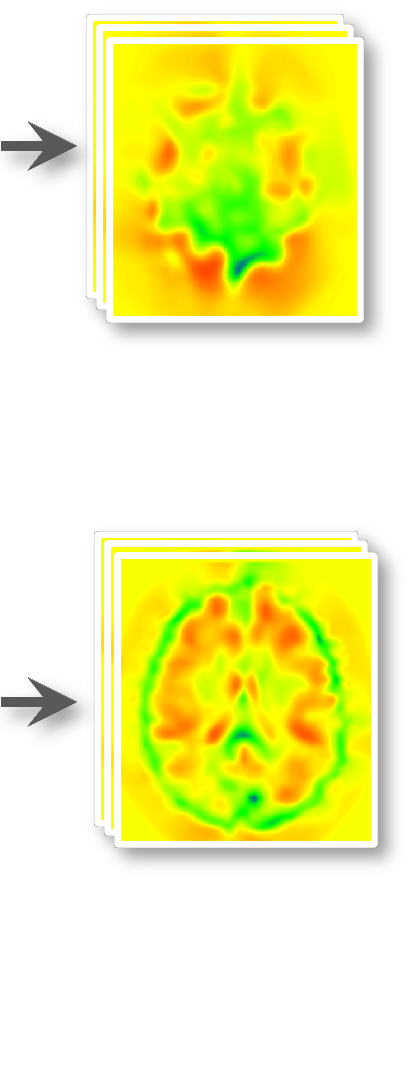}}}
  \caption{Our preprocessing pipeline starts with the (a) raw CT (upper) and raw MRI (lower) images. Next, (b) the CT scans go through contrast stretching, and the MRI scans go through bias correction. Then, (c) Brain Extraction Tool (BET) is applied to both images to remove non-brain areas. Following that, (d) Multi-stage registration involves aligning all subjects together, transforming the images to the MNI152 coordinate space, and then using MRI to perform per-subject registration of CT to MRI. Finally, given the deformable transformations from (d), we compute the Jacobian maps in (e).}
  \label{preprocessing_pipeline}
  \end{figure*}

\section{Methods}

\subsection{Data}
For the purpose of multimodality data, we use the recently released Open Access Series of Imaging Studies (OASIS)-3 dataset \cite{lamontagne2019oasis} which is composed of MRI, PET, and CT scan images from 1377 participants. The participants consist of 755 cognitively normal adults and 622 individuals at different stages of cognitive decline, spanning a wide age range from 42 to 95 years. CT imaging can reveal the atrophy or shrinkage of specific brain regions, which can indicate dementia including AD \cite{hhs2022biomarkers}. MRI provides detailed images of body structure and reveals progressive cerebral atrophy, which is best shown with T1w volumetric sequences. We did not select PET data because of its poor quality in terms of spatial resolution; in fact, MRI and CT are commonly used in structural assessments of brain atrophy whereas PET is less relevant.

Clinical assessments and diagnosis of the scans were based on the clinical dementia rating (CDR) scores of the participants, where a CDR score of 0 indicates no dementia, and CDR scores of 0.5, 1, 2, and 3 represent very mild, mild, moderate, and severe dementia, respectively \cite{marcus2010open}. Very mild can also be referred to as the MCI stage of AD \cite{marcus2010open}. Hence, we combined the mild to moderate participants to create four classes of AD dementia classification: normal, MCI, mild AD, and severe AD. These four classes constitute a more informative granularity as compared to classic binary classification.

\subsection{Preprocessing}
Preprocessing is imperative to ensure accurate classification of heterogeneous medical data as present in our study. We design a preprocessing pipeline as shown in \fref{preprocessing_pipeline}, which consists of three steps.

\subsubsection{Bias correction and contrast stretching}
For MRI images, we apply \textit{bias field correction} 
to reduce the spatially varying intensity bias, which could occur due to factors such as magnetic field inhomogeneities and acquisition artifacts. We perform bias field correction using FMRIB's Linear Image Registration Tool (FLIRT) \cite{jenkinson2002improved}. On the other hand, for CT images, we apply \textit{contrast stretching} to improve the visual perception and diagnostic value of CT images. Contrast stretching involves rescaling the pixel intensities to exploit the full dynamic range of the display, for which we take the contrast stretching portion out of the framework suggested by Kuijf et al. \cite{kuijf2013registration}. After that, we perform brain extraction using the Brain Extraction Tool (BET) \cite{smith2002fast} for both MRI and CT images to remove non-brain portion.


\subsubsection{Multi-stage registration}
This second step is a process of aligning and transforming images into a common coordination system. In a nutshell, we align the MRI brain images of all subjects together and transform them to the Montreal Neurological Institute's 152 brain template (MNI152), a standardized anatomical brain template widely adopted in neuroimaging research. This transformation involves mapping the images to a common coordinate system, allowing for meaningful comparison and analysis across different subjects. Next, to achieve alignment across modalities, we use the registered MRI images as the reference and align CT images to them. However, considering the unique shape of each subject's brain, we seek \textit{per-subject registration} which does not exist in prior work. To do so, we align each CT scan to its corresponding subject's MRI scan instead of a standard MRI brain template in order to account for individual differences. 

To elaborate, in aligning MRI, we first apply {\em deformable image registration} (DIR) to T1-weighted (T1w) MRI images to map them to MNI152. DIR is a computational technique used to align two images by deforming one image to match the shape and appearance of the other. Unlike rigid image registration (RIR), which involves only translation, rotation, and scaling \cite{oh2017deformable}, we apply DIR because we want to compute the Jacobian determinant (\sref{sec:jacob}) which needs a deformable transformation.  To this end, we look for a transformation \textbf{T} that aligns a {\em moving image} $M$, which is the source image to be transformed, to a {\em fixed image} $F$ which is the reference or target image. The transformation is represented as a function that deforms the source image to minimize the discrepancy between the transformed moving image $\mathbf{T}(M)$ and the fixed image $F$, and is typically a linear combination in the form of
\begin{equation}
    \mathbf{T}(\mathbf{x}) = \mathbf{x} + \mathbf{D}(\mathbf{x})
\end{equation}
where $\mathbf{x}$ represents the spatial location of any voxel 
in the input image, and $\mathbf{D}(\cdot)$ represents a displacement or deformation operation, obtained from the following optimization:
\begin{equation}
 \mathbf{D}(\mathbf{x}) =\arg \min_{\mathbf{D}}\left( \operatorname{sim}(F,\, \mathbf{T}(M))+\alpha \operatorname{reg}(\mathbf{T}) \right)
\end{equation}
where $\operatorname{sim}$ is a similarity metric, and $\operatorname{reg}$ is a regularization term that encourages smoothness in the deformation operation $\mathbf{D}$ to avoid overly complex or noisy deformations.



\begin{figure*}
\centering
\subfloat[Multimodal data are early fused depth-wise and fed to a 3D CNN.]{ \includegraphics[width=0.82\textwidth]{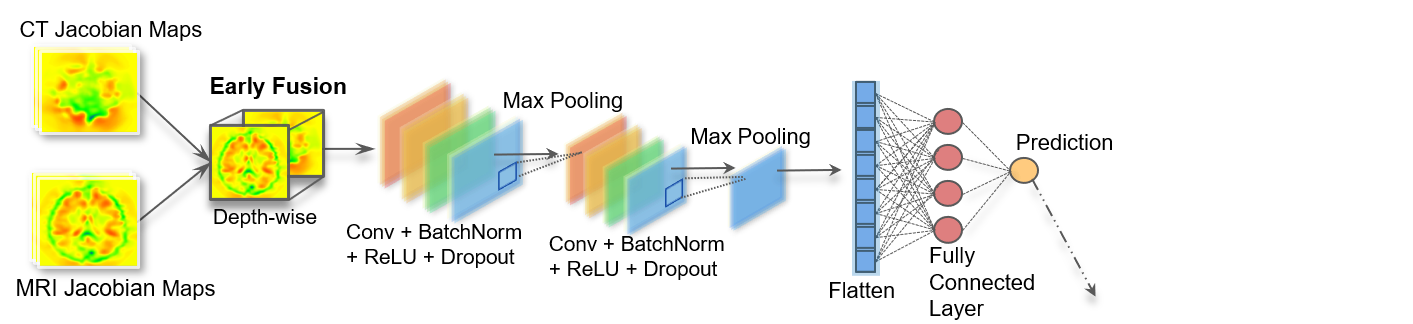}}
 \vspace{-0.1cm}
  \hfill
  \subfloat[Single-modal data is handled by RF through a deep feature extractor.]{\includegraphics[width=0.82\textwidth]{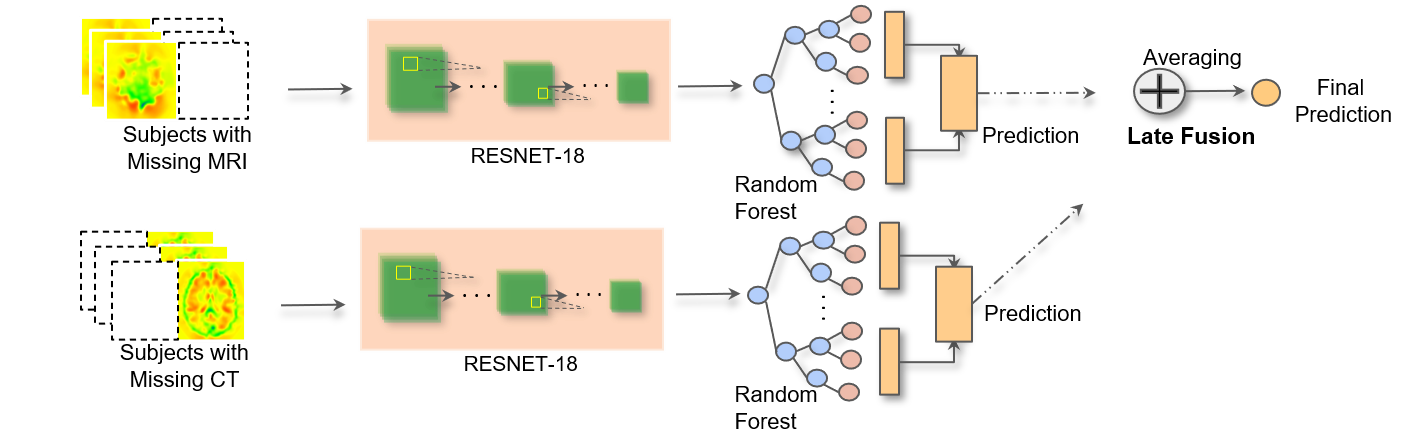}}
   
  \caption{The proposed Early-Late Fusion framework that follows after our preprocessing pipeline.} 
  \label{fig:elf_arch}
\end{figure*}

Next, we align each CT scan of each subject to its corresponding preregistered T1w MRI image using DIR, as we want to transform CT to the MNI152 coordinate space and compute the CT's Jacobian determinant. For CT subjects with missing T1w MRI scans, we align them with the T1w MRI of {\em another} subject that has the same label (one of the 4 AD classes) and the most similar skewness and kurtosis as this subject (more details in \sref{proposed_architecure}).  

\subsubsection{Transforming to Jacobian Domain}\label{sec:jacob}
In the last step, we transform both MRI and CT scans to the Jacobian domain (JD). This step was inspired by the ability of Jacobian determinants to shed light on local brain morphometry and shape changes, which was demonstrated by Abbas et al. \cite{abbas2023transformed}. The transformation function $\mathbf{T}(\mathbf{x})$ is computed at each voxel $\mathbf{x}$ with respect to its coordinates $(i, j, k)$. The first-order partial derivative of  $\mathbf{T}(\mathbf{x})$ with respect to the transformed coordinates $(i_\delta, j_\delta, k_\delta)$ forms the Jacobian matrix $J(\mathbf{x})$.
\begin{equation}
J(\mathbf{x})=\frac{\partial \mathbf{T}(\mathbf{x})}{\partial \mathbf{x}} = \left[\frac{\partial \mathbf{T}(\mathbf{x}) }{\partial i_\delta} \frac{\partial \mathbf{T}(\mathbf{x})}{\partial j_\delta}   \frac{\partial \mathbf{T}(\mathbf{x})}{\partial k_\delta}            \right]
\end{equation}

By computing the determinant of $J(\mathbf{x})$ for each voxel, the {\em Jacobian map}, we get the types of deformations $\mathbf{D}$  for each voxel $x$ resulting from the registration 
\begin{equation}
\text{type of }\mathbf{D} = \begin{cases}
\text{{volume compression}} & \text{{if }} |J(\mathbf{x})| < 1 \\
\text{{volume expansion}} & \text{{if }} |J(\mathbf{x})| > 1 \\
\text{{no change}} & \text{{if }} |J(\mathbf{x})| = 1
\end{cases}
\end{equation}
The value of $|J(\mathbf{x})|$ identifies the brain's volume change at the voxel level and explains our motivation for employing JD features. We perform both registration and Jacobian using Advanced Normalization Tools (ANTs) \cite{avants2009advanced}.

%

\subsection{Ensemble fusion and classification framework} \label{proposed_architecure}

We design an early-late fusion architecture that uses the Jacobian maps obtained from our preprocessing pipeline as the input. As shown in \fref{fig:elf_arch}, we first perform an early fusion that concatenates all the Jacobian maps of each subject along the depth dimension, where each map is obtained from one modality (MRI or CT in our case) of the same subject. Thus for each subject, we obtain a 3D representation which is then fed to a 3D CNN model to learn and extract discriminative features. Unlike Liu et al. \cite{liu2018multi}, our 3D CNN incorporates batch normalization and dropout regularization to avoid overfitting, and ReLU instead of Tanh activation. We choose a dropout rate of 0.2, non-strided convolution kernel (3,3,3), and max pooling kernel (2,2,2) with stride (2,2,2), where the sizes of conv and pooling are adopted from \cite{liu2018multi}.
To accommodate small datasets, we keep our model lightweight by having only two convolution layers before passing the learned 3D feature map to a flattening layer, which converts the 3D map into a 1D vector. The vector is then passed to a fully connected layer with softmax which produces probability scores for each of our 4 AD classes.

To handle subjects with missing modalities, we pass each available scan, either MRI or CT in our case, of a subject $S_n$, to a random forest (RF) model for MRI or an RF model for CT, correspondingly. However, as RF does not work well on raw or transformed images directly, we add ResNet \cite{he2016deep} as a deep feature extractor before RF to train each RF, as such auto-extracted features have been shown to perform better than hand-engineered features \cite{orenstein2017transfer}. Note that our ResNet is pretrained (using ImageNet-1K) and hence no extra (and large) datasets are required.

To aggregate the predictions of the three models, CNN, RF-MRI, and RF-CT, all subjects' data should be given as input to all models. However, we need to reconcile the inhomogeneity between subjects who have missing modalities and those who have both modalities. To handle the former, we impute the missing modalities using hot deck imputation (HDI), which replaces missing values by drawing from an estimated distribution, typically the distribution of the available sample data \cite{myrtveit2001analyzing}. A simple form of HDI imputes the missing value by randomly selecting a value from the dataset. In this work, to enhance the training phase, we impute the values of a missing modality of a subject by borrowing values from a similar subject who has a more complete set of data with similar statistical characteristics in terms of kurtosis and skewness. That is, for a subject $S_n$ with modality $M_n$ and missing modality $C_n$, we replace $C_n$ with $C_m$ of another subject $S_m$ who has the same label (e.g. mild AD) and whose $M_m$ exhibits the closest kurtosis and skewness to $M_n$. To handle the latter, for the subjects with both modalities, we feed MRI and CT individually to RF-MRI and RF-CT, respectively.

Finally, we aggregate the three models' decisions by averaging the prediction probabilities of the CNN, RF-MRI, and RF-CT models. Specifically, given the three models' respective prediction probabilities over all the four classes, $\mathbf{P}_1(y_i|s)|_{i=1}^4$, $\mathbf{P}_2(y_i|s)|_{i=1}^4$, and $\mathbf{P}_3(y_i|s)|_{i=1}^4$, where $y_i$ is the class label for subject $s$, the aggregated prediction, $\mathbf{P}_{aggr}(y_i|s)$, is computed as:
\begin{equation}
\mathbf{P}_{aggr}(y_i|s) = \frac{1}{3} \Big( \mathbf{P}_1(y_i|s) + \mathbf{P}_2(y_i|s) + \mathbf{P}_3(y_i|s) \Big)
\end{equation}
and the class index is determined by $\arg\max_i\mathbf{P}_{aggr}(y_i|s)$.

\subsection{Extension to More Heterogeneous Modalities}
Our proposed ELF framework can be easily extended to more than two and more heterogeneous modalities. For image-based modalities such as MRI, CT, and PET, they simply go through the same proposed framework as in \fref{fig:elf_arch}, where the number of RF models will equal the number of modalities. For non-image-based modalities generated by non-vision sensors such as EEG, which provides electrical brain activity data \cite{mustafa2023brain}, and MEG, which captures magnetic field information, they will be passed to our RF models (\fref{fig:elf_arch}b) without ResNet. When there are missing modalities, the HDI-based technique described in \sref{proposed_architecure} still applies.

\section{Performance Evaluation}

{\bf Splitting data with overfitting avoidance.}
In the OASIS-3 dataset, each subject originally underwent multiple MRI and CT scan sessions at different points in time. Dividing subjects properly among training, validation, and testing tests is crucial when attempting to detect a specific pathology through patients' scans. Having the same subjects in multiple sets can lead to overfitting of the model to those subjects, and thus performing poorly on unseen subjects \cite{altay2021preclinical}. For this reason, for each subject we randomly chose only one MRI session and one CT scan session whose timestamp is the closest to the chosen MRI session, in order to avoid having the same subject in more than one set, thus creating two distinct sets of subjects (one comprising 80\% for training and the other 20\% for testing). For validation, we employed a stratified 10-fold cross-validation setup using the training data.

{\bf Handling class imbalance.}
There is an inherent class imbalance problem in the four AD classes since the majority of subjects are normal. To handle this, we use the Adaptive Synthetic (ADASYN) \cite{he2008adasyn} oversampling algorithm to generate synthetic samples for minority classes in the training phase. In addition, we use a class weighting strategy to prioritize the \textit{underrepresented} classes to avoid bias towards majority classes. In that strategy, we define a weight $w(c)$ for each class $c$ as the inverse of its frequency $freq(c)$ in the training set ($y_{train}$ denoting labels), and normalize the class weights as shown below:
\begin{equation}
\begin{array}{l}
\text{\small freq($c$)} = \operatorname{count}\left(y_{train}=c\right), \text{for } c \in C \vspace{0.2cm}
\\
\text{\small Inverse Class Frequencies:} \quad w(c) = \frac{1}{\text{freq}(c)},  \text{for } c \in C \vspace{0.2cm}
\\ 
\text{\small Normalized Class Weights:} \quad w_{\text{normalized}}(c) = \frac{w(c)}{\sum_{c \in C} w(c)}
\end{array}
\end{equation}

\begin{table*}[!t]
\centering
\caption{Comparison with reported state-of-the-art using OASIS-3 dataset for AD classification}
\begin{tblr}{
  colspec = {@{}Q[l,m,3cm] Q[l,m,2.5cm] Q[l,m,3cm] Q[l,m,1.9cm] Q[l,m,1.8cm] Q[l,m,1.7cm]@{}},
  hline{1-2,6} = {-}{},
}
\textbf{Model}       & \textbf{Modalities}  & \textbf{Classes}  & \textbf{Sensitivity (\%)} & \textbf{Specificity (\%)} & \textbf{Accuracy (\%)} \\
Salami et al. \cite{salami2022designing}           & MRI (T1w)         & AD, CN        & 86.01    & 85.04       & 87.75 \\
Massalimova et al. \cite{massalimova2021input}    & MRI (T1w, DTI)     & NC, MCI, AD   & 96      & {\bf 96}          & 96  \\
Lazli et al. \cite{lazli2019computer}             & MRI, PET           & AD, healthy    &  92.00    &  91.78      & 91.46  \\
\textbf{ELF (ours)} & MRI (T1w), CT       & Normal, MCI, mild AD, severe AD          & \textbf{97.19}      & 95.19         & \textbf{98.76}
\end{tblr}
\label{table2}
\end{table*}

The normalized class weights are employed in the training phase, so the model is encouraged to pay more attention to the underrepresented classes, thereby mitigating bias towards the majority classes.

{\bf Performance metrics.}
We measure the performance of our four-class classification problem in terms of specificity and sensitivity, in addition to classification accuracy. We measure the sensitivity and specificity for each class and take the average. Sensitivity, also referred to as \textit{true positive rate} (TPR), is the ratio of correct positive predictions to the total number of actual positive samples, for class $i$. Specificity, also known as \textit{true negative rate} (TNR), is the ratio of correct negative predictions to the total number of actual negative samples, for all classes except $i$. In simpler words, TNR is the probability that an actual negative will test negative for class $i$. TPR and TNR can be represented as: 
\begin{align}
\hspace{-2mm}\mathrm{TPR(i)} = \frac{\mathrm{TP(i)}}{\mathrm{TP(i)}+\mathrm{FN(i)}},
    \mathrm{TNR(i)} = \frac{\mathrm{TN(\urcorner i)}}{\mathrm{TN(\urcorner i)}+\mathrm{FP(\urcorner i)}}
\end{align}
where $TP$, $FN$, $TN$, and $FP$ represent true positive, false negative, true negative, and false positive, respectively, and $\urcorner i$ represents the exclusion of class $i$.

{\bf Comparison with the state-of-the-art.} We first compare the results of our proposed ELF with the results reported in the recent papers that also used OASIS-3 dataset. \tref{table2} shows the results achieved by different studies in terms of sensitivity, specificity, and accuracy for the task of classifying individuals into different classes. The number of classes and modalities used in a model can impact its performance. Having more classes requires a more accurate discrimination ability between multiple categories. In addition, the model can capture more aspects of the disease by using more modalities, potentially improving its predictive power. Our proposed ELF performs the best in terms of accuracy (97.19\%) and is on par with \cite{massalimova2021input} in terms of specificity. Notably, our method tackles the task of classifying AD into four classes, which is harder than the binary classification problems addressed in other research papers. Salami et al. \cite{salami2022designing}  used a single modality (MRI-T1w) and classified AD into two classes, achieving an accuracy of 87.75\%.  Lazli et al. \cite{lazli2019computer}  achieved a slightly higher performance (91.46\%) in classifying AD into two classes using two modalities (MRI and PET). On the other hand, Massalimova et al. \cite{massalimova2021input} classified AD into three classes using two types of MRI (T1w and DTI), achieving an accuracy of 96\%.

\begin{table}[!t]
\centering
\caption{Ablation Study}
\begin{tblr}{
  hline{1-2,6} = {-}{},
}
\textbf{Model} & \textbf{Accuracy (\%)} & \textbf{Sensitivity (\%)} & \textbf{Specificity (\%)}  \\
CNN      & 91.02      & 83.37          & 87.21         \\
RF CT     & 94.26     & 86.79         & 90.52        \\
RF MRI     & 89.35       &  83.14         & 94.47         \\
\textbf{ELF} & \textbf{97.19}      & {\bf 95.19}         & \textbf{98.76}             
\end{tblr}
\label{table:Results}
\end{table}
\begin{figure*}
\begin{subfigure}[b]{0.34\textwidth}
         \centering
         \includegraphics[width=1\linewidth,height=4.75cm]{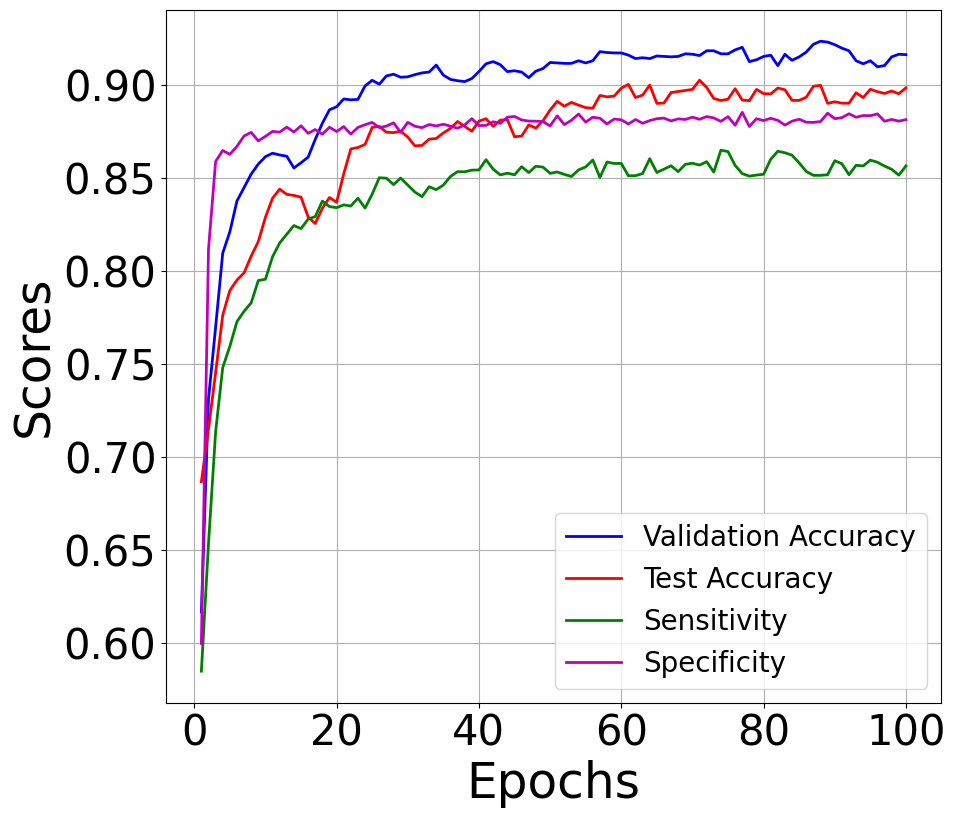}
\caption{\centering CNN training over Early fused multimodal data.}\label{subplot1}
     \end{subfigure}
\hfill    
\begin{subfigure}[b]{0.32\textwidth}
         \centering
        \includegraphics[width=1\linewidth]{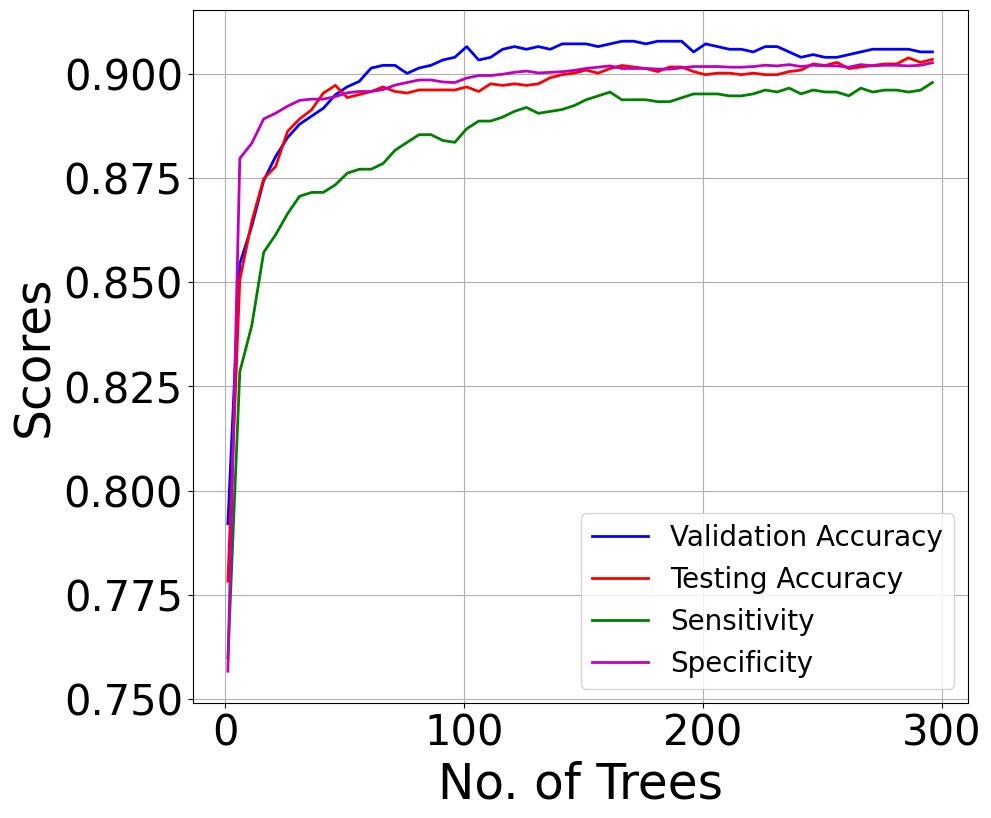}
   \caption{\centering  RF training over CT images.}
  \label{subplot2}
     \end{subfigure}
     \hfill
\begin{subfigure}[b]{0.31\textwidth}
         \centering
      \includegraphics[width=1\linewidth]{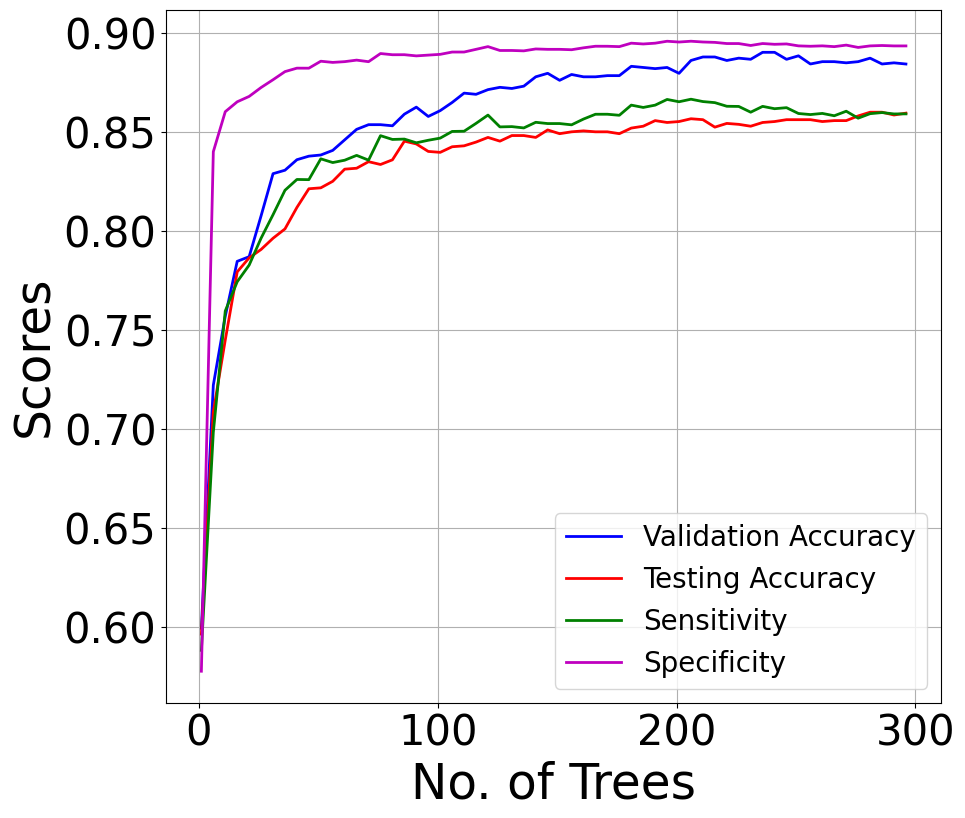}
    \caption{\centering RF training over MRI images.}
    \label{subplot3}
     \end{subfigure}
 \caption{Learning curves for CNN, RF-CT, and RF-MRI.}
   \label{Images/plots}
\end{figure*}

\begin{figure}[ht]
  \centering
 \includegraphics[width=0.75\linewidth,height=5cm]{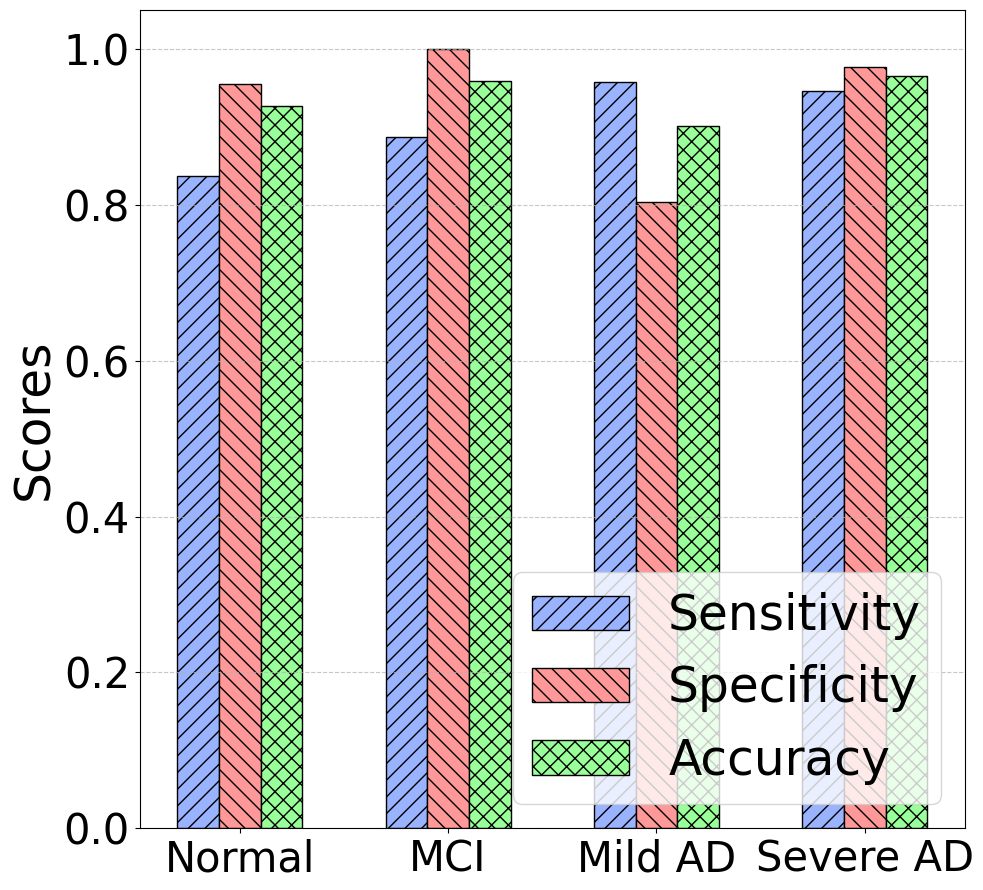}
  \caption{Testing Performance of the ELF model for each class}
  \label{bar_graph}

\end{figure}
{\bf In-depth investigation of the ELF framework.}
We trained the CNN model using the fused CT and MRI images and validated the performance using stratified 10-fold cross-validation. Similarly, we trained each random forest individually on the CT and MRI images and validated the results using stratified 10-fold cross-validation. \fref{Images/plots} shows the learning curves of each individual model. We ran the CNN model for 100 epochs on each fold and observed in \fref{subplot1} that the model reaches stability around epoch 40. The RF-CT and RF-MRI models were tested with varying numbers of trees, and in \fref{subplot2} we observed that the models perform well with fewer than 50 trees. Additionally, \fref{subplot3} shows that RF-MRI performs well after 50 epochs and remains stable throughout the rest of the curve. These curves clearly depict the effectiveness and efficiency of all three models. The learning rate and batch size were set to 0.001 and 4, respectively. 

\tref{table:Results} shows an ablation study on the three models. Our individual models have demonstrated strong performance, confirming our choice to use deep learning for fused MRI and CT, and RF for individual modalities. However, the ELF model achieves the highest accuracy (97.19\%), sensitivity (95.19\%), and specificity (98.76\%), implying that it's better at identifying negative and positive samples compared to the individual models. This suggests that aggregating the predictions of the three base models leads to improved performance and better overall results in classifying AD.

Finally, to ensure robust evaluation, the whole framework is evaluated on the test data, which was set aside using a stratified 80-20 train-test split of the dataset. Stratifying the data was crucial to make sure that both sets have the same data distribution across all four classes. \fref{bar_graph} shows the testing results for each class and demonstrates that the performance is good across all classes without being biased towards any particular one, further confirming the enhanced performance achieved through our proposed ELF.

\section{Conclusion}
In this paper, we introduced the Early-Late Fusion (ELF) approach to enhance the diagnosis of Alzheimer's disease across four distinct stages: normal, MCI, mild AD, and severe AD. We first provide a robust preprocessing pipeline that encompasses (1) per-subject registration which ensures alignment across different modalities of each subject, and (2) Jacobian domain transformation which empowers feature extraction in AD detection by furnishing information about brain morphometry and shape changes.

To handle the heterogeneous data modalities, our ELF framework incorporates both CNN and RF models to facilitate representation learning and classification, and leverages early fusion, late fusion, and an HDI technique. In our extensive experiments, ELF achieved a remarkable accuracy of 97.19\%, surpassing the most recent results reported in the literature. Furthermore, the ELF framework can be extended to even more heterogeneous modalities. Our research paves the way for more effective interventions and treatments of Alzheimer's disease diagnosis in the future.


%
%
%
{\small \bibliographystyle{ieeetran}
\bibliography{mybibliography}}

\end{document}